\documentclass{article}

\PassOptionsToPackage{square,sort,comma,numbers}{natbib}

\usepackage[final]{nips_2018}

\usepackage[utf8]{inputenc} 
\usepackage[T1]{fontenc}    
\usepackage{hyperref}       
\usepackage{url}            
\usepackage{booktabs}       
\usepackage{amsfonts}       
\usepackage{nicefrac}       
\usepackage{microtype}      
\usepackage{graphicx}      
\graphicspath{ {./} }

\title{Question Type Classification Methods Comparison}

\author{
  Tamirlan Seidakhmetov \\
  Stanford University\\
  \texttt{tamirlan@stanford.edu} \\
}

\begin{document}

\maketitle

\begin{abstract}
  The paper presents a comparative study of state-of-the-art approaches for question classification task: Logistic Regression, Convolutional Neural Networks (CNN), Long Short-Term Memory Network (LSTM) and Quasi-Recurrent Neural Networks (QRNN). All models use pre-trained GLoVe word embeddings and trained on human-labeled data. The best accuracy is achieved using CNN model with five convolutional layers and various kernel sizes stacked in parallel, followed by one fully connected layer. The model reached 90.7\% accuracy on TREC 10 test set. All the model architectures in this paper were developed from scratch on PyTorch, in few cases based on reliable open-source implementation.
\end{abstract}

\section{Introduction}
Text classification is a popular task in NLP which has a wide range of applications, such as document classification and question classification task. The latter task is an essential part of general question-answering and named entity recognition algorithms. With the recent progress of the artificial intelligence, tremendous progress has been made in this field, as when a question is asked, it is crucial to understand what the question is about before proceeding to answer search.

This paper focuses on short questions classification task and provides a comparative study on the state-of-the-art approaches in this field, such as QRNN, CNN, and LSTM models. All models are trained and tested on TREC 10 dataset.

\section{Related work}
Multiple attempts have been made to design various deep learning methods for text classification tasks and compare their performances on various questions. Le-Hong and Le in their work compared Convolutional Neural Networks (CNN), Recurrent Neural Networks (RNN),  Long Short-Term Memory Network (LSTM) and Feed-Forward Neural Network (FNN) on two datasets: UIUC English question classification dataset with fine-grained 50 classes and Vietnamese sentences from the vnExpress online newspaper \cite{DBLP:journals/corr/abs-1810-01656}. While paper gives good insights on how performances of different architectures compared, it did not consider novel Quasi-Recurrent Neural Network (QRNN) architecture.

The original QRNN paper also tried to compare their model with different variations of RNN, LSTM and CNN architectures. They used the IMDB movie review dataset to see how QRNN performs against other state-of-the-art models. One of the reported advantages of QRNN is that it performs well on longer text, however, researchers did not publish any results on QRNN's performance on shorter sentence classification \cite{DBLP:journals/corr/BradburyMXS16}.

\section{Approach}
\subsection{Data pre-processing}
Firstly, all individual words and punctuation like question marks and commas in the question sentence were tokenized into separate entities.
$$Xword = [token1, token2, . . . ,token N] \in Z^N$$
where $N$ is a number of tokens in the sentence.

After that, the words have been converted to word indices and then further converted to word embeddings using GLoVe pre-trained word vectors. The GLoVe vectors were pre-trained using 840 billion tokens from Common Crawl, and each token is mapped into a 300-dimensional vector \cite{pennington2014glove}. 
$$Xembeddings = GloveEmbedding(Xword) \in R^{NxD_{word}}$$
where $D_{word}$ is a number of dimensions of a word vector. 
Using these input features various models have been built.

\subsection{Logistic Regression}
As a baseline, a multi-class logistic regression model has been built. After each word was converted to $D_{word}=300$ dimensional word vector, the input became $NxD_{word}$, where $N$ is a number of words in a sentence. Then, the average pooling was applied per each sentence along each dimension, so that the output became a $1xD_{word}$ dimensional vector. 
$$Xavg = AveragePool(Xembeddings) \in R^{D_{word}}$$

Further, one layer of linear network has been applied followed by a Sigmoid function, which is equivalent to logistic regression. 
$$Xprob = Sigmoid(LinearLayer(Xavg)) \in R^C$$
The output is a C-dimensional output, where C is a number of classes that we are trying to predict, and the output value is a probability of each class. The code for the baseline model was implemented from scratch.

\subsection{Convolutional Neural Network}
Convolutional Neural Network (CNN) is a popular building block for many neural networks and widely used in image processing. Here CNN was implemented on the question classification task based on Yoon Kim’s “Convolutional Neural Networks for Sentence Classification” research work \cite{DBLP:journals/corr/Kim14f}. After the words have been converted to word embeddings, a series of convolution layers have been applied in parallel, with various kernel sizes $K$. Different kernel sizes correspond to various n-grams, so when kernel size is 2, the model looks for 2-grams, therefore learning a lower term meaning of words, whereas, for kernel size 5, CNN learns 5-grams which have higher term information about the question. The number of kernels, as well as their sizes, are important hyperparameters, that were tuned to find an optimal setting. 
$$Xconv\_k = Conv(Xembeddings), k \in K$$

Further, the outputs of all convolution layers with different kernel sizes are passed through the max-pooling step over time.
$$Xmaxpool\_k = MaxPool(Xconv\_k), k \in K$$

Then the results are stacked together. 
$$Xmaxpool\_all = Stack(Xmaxpool\_k), k \in K$$

During training time, the dropout layer is implemented for regularization. This step is omitted at inference time.
$$Xmaxpool\_all = Dropout(Xmaxpool\_all)$$

Finally, the output is passed through the fully connected layer and then a Softmax function. The output vector represents the probability scores of each output class.
$$Xlinear = LinearLayer(Xmaxpool\_all)$$
$$Xprob = Softmax(Xlinear)$$

Several experiments have been conducted to improve the original implementation of the CNN architecture in Yoon Kim's paper, specifically multiple fully connected layers were added at the end, to increase the complexity of the model and improve the accuracy. So the last two layers looked like the following:
$$Xlinear1 = LinearLayer(Xmaxpool\_all)$$
$$Xlinear2 = LinearLayer(Xlinear1)$$
$$Xlinear3 = LinearLayer(Xlinear2)$$
$$Xprob = Softmax(Xlinear3)$$

Each subsequent linear layer decreased the output size by half, except for the last one, which still has an output size equal to the number of classes that the model needs to predict.

The code for this part was inspired by similar open-source implementation of the CNN model \cite{github_cnn}, along with modifications in neural network architecture and hyperparameter tuning.

\begin{figure}[h]
\centering
\includegraphics[scale=0.41]{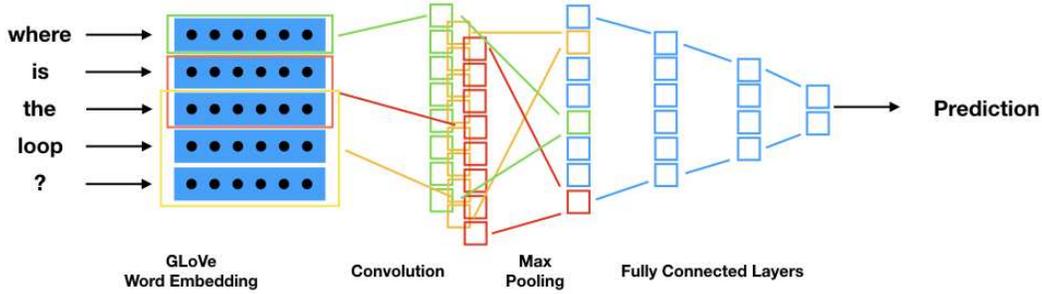}
\caption{CNN Architecture with 3 FC layers}
\end{figure}

\subsection{Long Short-Term Memory Network}
Long Short-Term Memory (LSTM) Network is a popular type of Recurrent Neural Networks (RNN), which became an essential building block for many neural network architectures. LSTM layers are frequently used in a bidirectional setup with multiple layers stacked together, to learn higher-level dependency of the words in a sentence. A similar work was done by Yangyang Shi and others, where they fed word embeddings into multiple stacked bi-LSTM layers and used the convolution layer across different bi-LSTM layers to capture features at different layers. Then the model followed by average pooling \cite{inproceedings}.

In our work, the convolutional layer is not used, as to test the capabilities of LSTM only model. The word embeddings are fed to multiple stacked bidirectional LSTM networks with dropout applied in between:
$$Xlstm_1 = BiLSTM_1(Xembeddings)$$
$$Xlstm_1 = Dropout(Xlstm_1)$$
$$Xlstm_2 = BiLSTM_2(Xlstm_1)$$
$$Xlstm_2 = Dropout(Xlstm_2)$$
$$. . .$$
$$Xlstm_n = BiLSTM_n(Xlstm_{n-1})$$
$$Xlstm_n = Dropout(Xlstm_n)$$

The product of stacked Bi-LSTM layers is passed through a fully connected linear layer, followed by a softmax function to obtain the prediction probability of each class.
$$Xprob = Softmax(LinearLayer(Xlstm_n))$$

The output is C-dimensional vector, where C is a number of classes

The code for this part is implemented from scratch.

\begin{figure}[h]
\centering
\includegraphics[scale=0.41]{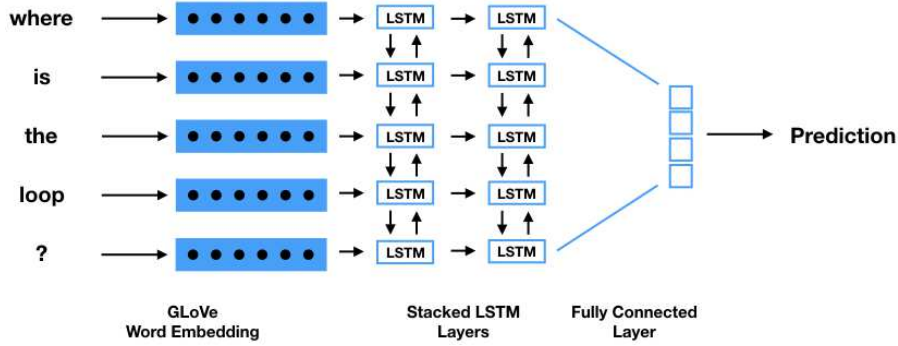}
\caption{LSTM Architecture with 2 stacked Bi-LSTM layers}
\end{figure}

\subsection{Quasi-Recurrent Neural Network}
Quasi-Recurrent Neural Networks is a relatively new approach for modeling a sequence data, introduced by James Bradbury and others at Salesforce in 2017 \cite{DBLP:journals/corr/BradburyMXS16}. QRNN is a novel model, which uses the advantages of previously well known two models: LSTM and CNN. While LSTM is good at capturing dependencies in sequential data for moderately long sequences, it frequently fails to perform well on very long sequences such as document and paragraph classification, or character-level models. Also, it has relatively slow performance, due to inability to do parallel computations. On the other hand, CNN allows faster computation, as convolution operations could be done in parallel, those allowing better scaling. QRNN tries to capture long term dependencies and allows parallelism, which is good for scaling \cite{DBLP:journals/corr/BradburyMXS16}.

QRNN consists of two sub-components: convolution and max-pooling, both of which are parallelizable. At a time step t, a convolution filter with filter width k is applied, starting from $x_{t-k+1}$ to $x_t$, mainly to be used at the next element in a sequence prediction tasks. A total of m convolution filters are applied as described above, followed by tanh nonlinearity. Forget and output gates with sigmoid functions are used at the pooling step.

Overall, the neural network architecture of QRNN model in this paper is similar to LSTM architecture in  previous section, except bidirectional LSTM is replaced by unidirectional QRNN layer with a convolution filter width 1 and 2:
$$Xqrnn_1 = QRNN_1(Xembeddings)$$
$$Xqrnn_1 = Dropout(Xqrnn_1)$$
$$Xqrnn_2 = QRNN_2(Xqrnn_1)$$
$$Xqrnn_2 = Dropout(Xqrnn_2)$$
$$Xprob = Softmax(LinearLayer(Xqrnn_2))$$

The code in this part was implemented from scratch, except the QRNN module itself, which has an official open-source implementation by the authors of the paper.

\newpage
\section{Experiments}
\subsection{Data}
TREC question classification dataset was used to compare the methods. It provides questions that mostly consist of one sentence and target is one of six classes associated with the question: Abbreviation, Entity, Description, Human, Location and Numeric. Here are the examples of questions in each class:

\begin{center}
Table 1: TREC dataset quesion-class examples
\end{center}
\begin{center}
 \begin{tabular}{||c c||} 
 \hline
 Question & Class \\ [0.5ex] 
 \hline\hline
 Who killed Gandhi? & Human \\ [0.5ex] 
 \hline
 What does the abbreviation AIDS stand for? & Abbreviation \\ [0.5ex] 
 \hline
 What do Mormons believe? & Description \\ [0.5ex] 
 \hline
 What is the date of Boxing Day? & Number \\ [0.5ex] 
 \hline
What is a female rabbit called? & Entity \\ [0.5ex]  
 \hline
Where is the highest point in Japan? & Location \\ [0.5ex]  
 \hline
\end{tabular}
\end{center}

There are a total of 5500 training examples and 500 test examples. The training set was further separated into 4500 training, 500 validation and 500 preliminary test examples.

\subsection{Evaluation method}
The model is being evaluated using an accuracy score, which is defined as:
$$Accuracy = \sum \frac {true\ labels} {all\ predicted\ labels}$$
True labels are correctly predicted values.

\subsection{Experimental details}
For data preparation, a torchtext text preprocessing library for PyTorch has been used. The library helps with building a word vocabulary, separate data into training, validation and test sets, convert tokenized sentences to indices, apply padding and build an iterator to iterate over batches. The iterator also separates sentences into batches by their lengths, so that shorter sentences appear next to shorter ones in the same batch, and minimal padding is applied. One deficiency of the library is that there is no way to apply padding to a sentence and have a predefined minimum sentence length, which is needed when CNN with larger kernel sizes is applied. For instance, when kernel size is 5, and maximum sentence length in a batch is 4, the model needs padding. Therefore padding and tokenization logic was implemented from scratch.

After the input pre-processing, GLoVe 300-dimensional pre-trained word vectors have been loaded to the Embedding layer. Also, Adam optimizer has been used at a training stage. The batch size is 64.

Both TREC and Books model training is done on GPU, due to training data sizes. All the models have been trained for 30-100 epochs, depending on the training data size.

The dropout rates used in the experiment range between 0.2-0.7 and the hidden layer size between 50 and 300. 

The training time of each model is around 20 min for the TREC dataset, and up to 10-15 hours for Books dataset on GPU.

\subsection{Results}
Table 2 shows a comparison of performances of CNN architectures with various kernel size combinations. It could be noted that as higher-level kernels are appended to the model, the accuracy gets better at the beginning and then plateaus: after reaching sizes 5 or 6, there is no much incremental increase. One possible reason for this is that many question sentences in the dataset are relatively short, and most of their meanings are captured by lower-level kernels. So two best CNN models with four (2,3,4,5) and five (2,3,4,5,6) kernel sizes are chosen to be further tested.

\begin{center}
Table 2: CNN kernel size comparison on TREC Internal Test Set
\end{center}
\begin{center}
 \begin{tabular}{||c c||} 
 \hline
 Model & Accuracy \\ [0.5ex] 
 \hline\hline
 CNN w kernels (2) & 85.8 \\ [0.5ex] 
 \hline
 CNN w kernels (2, 3) & 87.2 \\ [0.5ex] 
 \hline
 CNN w kernels (2, 3, 4) & 87.6 \\ [0.5ex] 
 \hline
 CNN w kernels (2, 3, 4, 5) & 88.8\\ [0.5ex] 
 \hline
 CNN w kernels (2, 3, 4, 5, 6) & 88.8\\ [0.5ex]  
 \hline
\end{tabular}
\end{center}

Table 3 below shows the comparison among different algorithms on TREC 10 and Books app test sets:


\begin{center}
Table 3: Test set comparison
\end{center}
\begin{center}
 \begin{tabular}{||c c||} 
 \hline
 Model & TREC 10 Accuracy \\ [0.5ex] 
 \hline\hline
 Logistic Regression & 87.3 \\ [0.5ex] 
 \hline
 Bi-LSTM - 2 Stacked Layers & 88.3 \\ [0.5ex]  
 \hline
 Bi-LSTM - 5 Stacked Layers & 82.4 \\ [0.5ex]  
 \hline
 CNN w kernels (2,3,4,5) + 1 FC Layer & 89.6 \\ [0.5ex] 
  \hline
 CNN w kernels (2,3,4,5,6) + 1 FC Layer & 90.7 \\ [0.5ex] 
 \hline
 CNN w kernels (2,3,4,5,6) + 3 FC Layers & 88.6 \\ [0.5ex] 
 \hline
 QRNN w 1 Layers and Window Size 1 & 77.8 \\ [0.5ex]
 \hline
 QRNN w 2 Layers and Window Size 1 & 86.2 \\ [0.5ex]
 \hline
 QRNN w 2 Layers and Window Size 2 & 88.0 \\ [0.5ex]
 \hline
\end{tabular}
\end{center}

Baseline Logistic Regression model shows very good performance on TREC 10 dataset, however, 2-layer stacked Bi-LSTM model performs a little better, improving a baseline result by 1\%.

CNN based approach with kernel sizes from 2 to 6 and one fully connected layer shows the best performance among all models, reaching 90.7\% accuracy.

More complex QRNN model does well on TREC 10, but simpler alternatives do not perform well on this dataset, trailing behind the baseline model. This might be due to a small training data size for this task.

\section{Analysis}
The models' comparison is shown in Table 3. Here are a few frequent errors that models do:
\begin{enumerate}
  \item Classify text with names as "human".
  \item Out-of-vocabulary words. There is a special embedding for all out-of-vocabulary words, it means the model sees all out-of-vocabulary words similarly. This decreases the performance of the models. The problem mostly affects unusual names and new or foreign words.
  \item Misspelling. This is a subset of an out-of-vocabulary word problem, however, this problem worth a separate bullet point, as people tend to misspell frequently.

\end{enumerate}
\newpage
\bibliography{FinalPaper_noApple.bbl}{}
\bibliographystyle{unsrtnat}
\end{document}